\documentclass[runningheads]{llncs}
\usepackage{amsmath}
\usepackage{graphicx}
\usepackage{bm}
\usepackage{multirow}
\usepackage{amssymb}
\usepackage{subfigure}
\usepackage[symbol]{footmisc}

\begin{document}
\title{Accurate Spectral Super-resolution from Single RGB Image Using Multi-scale CNN}
%
%
\author{
Yiqi Yan\inst{1} \and
Lei Zhang\inst{2} \and
Jun Li\inst{4} \and
Wei Wei\inst{2,3}\thanks{Corresponding should be address to W. Wei.} \and
Yanning Zhang\inst{2,3}
}

\authorrunning{
Yiqi Yan et al.
}
%
\institute{
School of Electronics and Information, Northwestern Polytechnical University, Xi'an, China \and
School of Computer Science, Northwestern Polytechnical University, Xi'an, China \and
National Engineering Laboratory for Integrated Aero-Space-Ground-Ocean Big Data Application Technology, China \and
Guangdong Provincial Key Laboratory of Urbanization and Geo-simulation, School of Geography and Planning, Sun Yat-Sen University, Guangzhou, China \\
\email{yanyiqinwpu@gmail.com}, \email{zhanglei211@mail.nwpu.edu.cn}, \\
\email{lijun48@mail.sysu.edu.cn}, \email{\{weiweinwpu,ynzhang\}@nwpu.edu.cn}
}
\maketitle              
\begin{abstract}
Different from traditional hyperspectral super-resolution approaches that focus on improving the spatial resolution, spectral super-resolution aims at producing a high-resolution hyperspectral image from the RGB observation with super-resolution in spectral domain. However, it is challenging to accurately reconstruct a high-dimensional continuous spectrum from three discrete intensity values at each pixel, since too much information is lost during the procedure where the latent hyperspectral image is downsampled (e.g., with $\times$10 scaling factor) in spectral domain to produce an RGB observation. To address this problem, we present a multi-scale deep convolutional neural network (CNN) to explicitly map the input RGB image into a hyperspectral image. Through symmetrically downsampling and upsampling the intermediate feature maps in a cascading paradigm, the local and non-local image information can be jointly encoded for spectral representation, ultimately improving the spectral reconstruction accuracy. Extensive experiments on a large hyperspectral dataset demonstrate the effectiveness of the proposed method.
\end{abstract}
\section{Introduction}

Hyperspectral imaging encodes the reflectance of the scene from hundreds or thousands of bands with a narrow wavelength interval (e.g., 10nm) into a hyperspectral image. Different from conventional images, each pixel in the hyperspectral image contains a continuous spectrum, thus allowing the acquisition of abundant spectral information. Such information has proven to be quite useful for distinguishing different materials. Therefore, hyperspectral images have been widely exploited to facilitate various applications in computer vision community, such as visual tracking~\cite{Reference1}, image segmentation~\cite{Reference2}, face recognition~\cite{Reference3}, scene classification~\cite{Reference6}, and anomaly detection~\cite{Reference8}.

The acquisition of spectral information, however, comes at the cost of decreasing the spatial resolution of hyperspectral images. This is because a fewer number of photons are captured by each detector due to the narrower width of the spectral bands. In order to maintain a reasonable signal-to-noise ratio (SNR), the instantaneous field of view (IFOV) needs to be increased, which renders it difficult to produce hyperspectral images with high spatial resolution. To address this problem, many efforts have been made for the hyperspectral imagery super-resolution.

Most of the existing methods mainly focus on enhancing the spatial resolution of the observed hyperspectral image. According to the input images, they can be divided into two categories: $(1)$ fusion based methods where a high-resolution conventional image ($e.g.$, RGB image) and a low-resolution hyperspectral image are fused together to produce a high-resolution hyperspectral image \cite{Reference50,Reference51} $(2)$ single image super-resolution which directly increases the spatial resolution of a hyperspectral image \cite{Reference72,Reference76,Reference77,Reference80}. Although these methods have shown effective performance, the acquisition of the input hyperspectral image often requires specialized hyperspectral sensors as well as extensive imaging cost. To mitigate this problem, some recent literature \cite{Reference20,Reference23,Reference27,Reference30} turn to investigate a novel hyperspectral imagery super-resolution scheme, termed spectral super-resolution, which aims at improving the spectral resolution of a given RGB image. Since the input image can be easily captured by conventional RGB sensors, imaging cost can be greatly reduced.

However, it is challenging to accurately reconstruct a hyperspectral image from a single RGB observation, since mapping three discrete intensity values to a continuous spectrum is a highly ill-posed linear inverse problem. To address this problem, we propose to learn a complicated non-linear mapping function for spectral super-resolution with deep convolutional neural networks (CNN). It has been shown that the 3-dimensional color vector for a specific pixel can be viewed as the downsampled observation of the corresponding spectrum. Moreover, for a candidate pixel, there often exist abundant locally and no-locally similar pixels ($i.e.$ exhibiting similar spectra) in the spatial domain. As a result, the color vectors corresponding to those similar pixels can be viewed as a group of downsampled observations of the latent spectra for the candidate pixel. Therefore, accurate spectral reconstruction requires to explicitly consider both the local and non-local information from the input RGB image. To this end, we develop a novel multi-scale CNN. Our method jointly encodes the local and non-local image information through symmetrically downsampling and upsampling the intermediate feature maps in a cascading paradigm, thus enhancing the spectral reconstruction accuracy. We experimentally show that the proposed method can be easily trained in an end-to-end scheme and beat several state-of-the-art methods on a large hyperspectral image dataset with respect to various evaluation metrics.

Our contributions are twofold:

\begin{itemize}
 \item We design a novel CNN architecture that is able to encode both local and non-local information for spectral reconstruction.
 \item We perform extensive experiments on a large hyperspectral dataset and obtain the state-of-the-art performance.
\end{itemize}

\section{Related Work}

This section gives a brief review of the existing spectral super-resolution methods, which can be divided into the following two categories.

\textbf{Statistic based methods} This line of research mainly focus on exploiting the inherent statistical distribution of the latent hyperspectral image as priors to guide the super-resolution~\cite{Reference78,Reference79}. Most of these methods involve building overcomplete dictionaries and learning sparse coding coefficients to linearly combine the dictionary atoms. For example, in~\cite{Reference20}, Arad $et$ $al.$ leveraged image priors to build a dictionary using K-SVD~\cite{Reference21}. At test time, orthogonal matching pursuit~\cite{Reference22} was used to compute a sparse representation of the input RGB image.~\cite{Reference23} proposed a new method inspired by A+~\cite{Reference25}, where sparse coefficients are computed by explicitly solving a sparse least square problem. These methods directly exploit the whole image to build image prior, ignoring local and non-local structure information. What's more, since the image prior is often handcrafted or heuristically designed with shallow structure, these methods fail to generalize well in practice.

\textbf{Learning based methods} These methods directly learn a certain mapping function from the RGB image to a corresponding hyperspectral image. For example,~\cite{Reference27} proposed a training based method using a radial basis function network. The input data was pre-processed with a white balancing function to alleviate the influence of different illumination. The total reconstruction accuracy is affected by the performance of this pre-processing stage. Recently, witnessing the great success of deep learning in many other ill-posed inverse problems such as image denoising~\cite{Reference28} and single image super-resolution~\cite{Reference29}, it is natural to consider using deep networks (especially convolutional neural networks) for spectral super-resolution. In~\cite{Reference30}, Galliani $et$ $al.$ exploited a variant of fully convolutional DenseNets (FC-DenseNets~\cite{Reference31}) for spectral super-resolution. However, this method is sensitive to the hyper-parameters and its performance can still be further improved.

\section{Proposed Method}

In this section, we will introduce the proposed multi-scale convolution neural network in details. Firstly, we introduce some building blocks which will be utilized in our network. Then, we will illustrate the architecture of the proposed network.

\subsection{Building Blocks}

\begin{table}[t]
\centering
\caption{Basic building blocks of our network}
\label{tab1}
\subtable{
\begin{tabular}{c}
\hline
\textbf{Double Conv}      \\ \hline
$3 \times 3$ convolution  \\
Batch normalization       \\
Leaky ReLU                \\
2D Dropout                \\ \hline
$3 \times 3$ convolution  \\
Batch normalization       \\
Leaky ReLU                \\
2D Dropout                \\ \hline
\end{tabular}
}
\subtable{
\begin{tabular}{c}
\hline
\textbf{Downsample}       \\ \hline
$2 \times 2$ max-pooling  \\ \hline
                          \\
                          \\
                          \\ \hline
\textbf{Upsample}         \\ \hline
Pixel shuffle             \\ \hline
\end{tabular}
}
\end{table}

There are three basic building blocks in our network. Their structures are shown in Table~\ref{tab1}.

\textbf{Double convolution (Double Conv) block} consists of two $3\times3$ convolutions. Each of them is followed by batch normalization, leaky ReLU and dropout. We exploit batch normalization and dropout to deal with overfitting.

\textbf{Downsample block} contains a regular max-pooling layer. It reduces the spatial size of the feature map and enlarges the receptive field of the network.

\textbf{Upsample block} is utilized to upsample the feature map in the spatial domain. To this end, much previous literature often adopts the transposed convolution. However, it is prone to generate checkboard artifacts. To address this problem, we use the pixel shuffle operation~\cite{Reference49}. It has been shown that pixel shuffle alleviates the checkboard artifacts. In addition, due to not introducing any learnable parameters, pixel shuffle also helps improve the robustness against over-fitting.

\begin{figure}[t]
\centering
\includegraphics[width=0.8\textwidth]{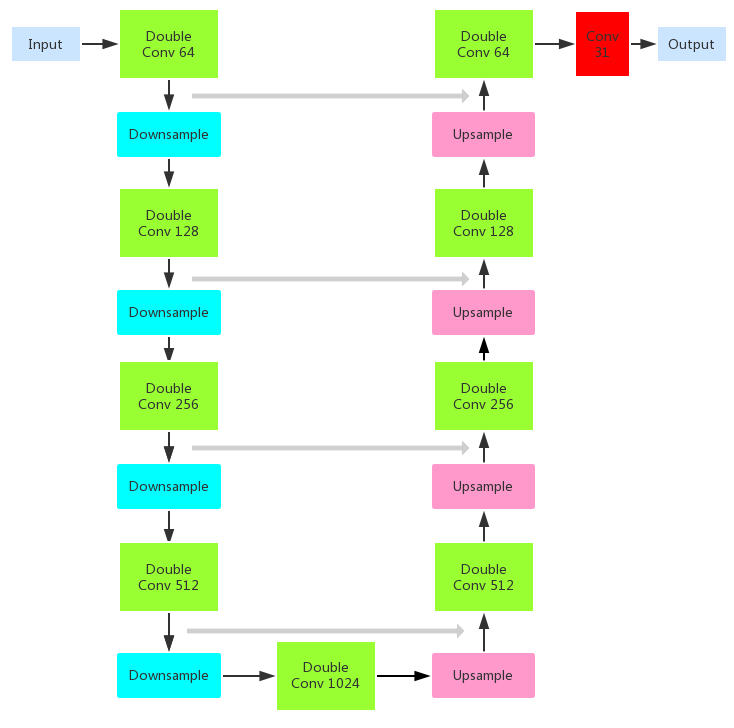}
\caption{Diagram of the proposed method. ``Conv $m$'' represents convolutional layers with an output of $m$ feature maps. We use $3 \times 3$ convolution in green blocks and $1 \times 1$ convolution in the red block. Gray arrows represent feature concatenation.}
\label{fig1}
\end{figure}

\subsection{Network Architecture}

Our method is inspired by the well known U-Net architecture for image segmentation~\cite{Reference43}. The overall architecture of the proposed multi-scale convolution neural network is depicted in Figure~\ref{fig1}. The network follows the encoder-decoder pattern. For the \textbf{encoder} part, each downsampling step consists of a ``Double Conv'' with a downsample block. The spatial size is progressively reduced, and the number of features is doubled at each step. The \textbf{decoder} is symmetric to the encoder path. Every step in the decoder path consists of an upsampling operation followed by a ``Double Conv'' block. The spatial size of the features is recovered, while the number of features is halved every step. Finally, a $1 \times 1$ convolution maps the output features to the reconstructed 31-channel hyperspectral image. In addition to the feedforward path, skip connections are used to concatenate the corresponding feature maps of the encoder and decoder.

Our method naturally fits the task of spectral reconstruction. The encoder can be interpreted as extracting features from RGB images. Through downsampling in a cascade way, the receptive field of the network is constantly increased, which allows the network to ``see'' more pixels in an increasingly larger field of view. By doing so, both the local and non-local information can be encoded to better represent the latent spectra. The symmetric decoder procedure is employed to reconstruct the latent hyperspectral images based on these deep and compact features. The skip connections with concatenations are essential for introducing multi-scale information and yielding better estimation of the spectra.

\section{Experiments}

\subsection{Datasets}

In this study, all experiments are performed on the NTIRE2018 dataset~\cite{Reference34}. This dataset is extended from the ICVL dataset~\cite{Reference20}. The ICVL dataset includes $203$ images captured using Specim PS Kappa DX4 hyperspectral camera. Each image is of size $1392\times1300$ in spatial resolution and contains $519$ spectral bands in the range of $400\sim1000 nm$. In experiments, $31$ successive bands ranging from $400\sim700 nm$ with $10 nm$ interval are extracted from each image for evaluation. In the NTIRE2018 challenge, this dataset is further extended by supplementing $53$ extra images of the same spatial and spectral resolution. As a result, $256$ high-resolution hyperspectral images are collected as the training data. In addition, another $5$ hyperspectral images are further introduced as the test set. In the NTIRE2018 dataset, the corresponding RGB rendition is also provided for each image. In the following, we will employ the RGB-hyperspectral image pairs to evaluate the proposed method.

\begin{table}[ht!]
\centering
\caption{Quantitative results on each test image.}
\label{tab4}
\resizebox{\textwidth}{!}{
\begin{tabular}{ccccccc}
\hline
\multicolumn{7}{c}{$RMSE_{1}$}                                                                                                  \\ \hline
                    & BGU\_00257      & BGU\_00259      & BGU\_00261      & BGU\_00263      & BGU\_00265      & Average         \\ \hline
Interpolation       & 1.8622          & 1.7198          & 2.8419          & 1.3657          & 1.9376          & 1.9454          \\
Arad $et$ $al.$     & 1.7930          & 1.4700          & 1.6592          & 1.8987          & 1.2559          & 1.6154          \\
A+                  & 1.3054          & 1.3572          & 1.3659          & 1.4884          & 0.9769          & 1.2988          \\
Galliani $et$ $al.$ & 0.7330          & 0.7922          & 0.8606          & 0.5786          & \textbf{0.8276} & 0.7584          \\
Our                 & \textbf{0.6172} & \textbf{0.6865} & \textbf{0.9425} & \textbf{0.5049} & 0.8375          & \textbf{0.7177} \\ \hline
\multicolumn{7}{c}{$RMSE_{2}$}                                                                                                  \\ \hline
                    & BGU\_00257      & BGU\_00259      & BGU\_00261      & BGU\_00263      & BGU\_00265      & Average         \\ \hline
Interpolation       & 3.0774          & 2.9878          & 4.1453          & 2.0874          & 3.9522          & 3.2500          \\
Arad $et$ $al.$     & 3.4618          & 2.3534          & 2.6236          & 2.5750          & 2.0169          & 2.6061          \\
A+                  & 2.1911          & 1.9572          & 1.9364          & 2.0488          & 1.3344          & 1.8936          \\
Galliani $et$ $al.$ & 1.2381          & \textbf{1.2077} & \textbf{1.2577} & 0.8381          & \textbf{1.6810} & \textbf{1.2445} \\
Ours                & \textbf{0.9768} & 1.3417          & 1.6035          & \textbf{0.7396} & 1.7879          & 1.2899          \\ \hline
\multicolumn{7}{c}{$rRMSE_{1}$}                                                                                                 \\ \hline
                    & BGU\_00257      & BGU\_00259      & BGU\_00261      & BGU\_00263      & BGU\_00265      & Average         \\ \hline
Interpolation       & 0.0658          & 0.0518          & 0.0732          & 0.0530          & 0.0612          & 0.0610          \\
Arad $et$ $al.$     & 0.0807          & 0.0627          & 0.0624          & 0.0662          & 0.0560          & 0.0656          \\
A+                  & 0.0580          & 0.0589          & 0.0612          & 0.0614          & 0.0457          & 0.0570          \\
Galliani $et$ $al.$ & 0.0261          & 0.0268          & 0.0254          & 0.0237          & 0.0289          & 0.0262          \\
Ours                & \textbf{0.0235} & \textbf{0.0216} & \textbf{0.0230} & \textbf{0.0205} & \textbf{0.0278} & \textbf{0.0233} \\ \hline
\multicolumn{7}{c}{$rRMSE_{2}$}                                                                                                 \\ \hline
                    & BGU\_00257      & BGU\_00259      & BGU\_00261      & BGU\_00263      & BGU\_00265      & Average         \\
Interpolation       & 0.1058          & 0.0933          & 0.1103          & 0.0759          & 0.1338          & 0.1038          \\
Arad $et$ $al.$     & 0.1172          & 0.0809          & 0.0819          & 0.0685          & 0.0733          & 0.0844          \\
A+                  & 0.0580          & 0.0589          & 0.0612          & 0.0614          & 0.0457          & 0.0610          \\
Galliani $et$ $al.$ & 0.0453          & \textbf{0.0372} & \textbf{0.0331} & 0.0317          & \textbf{0.0562} & \textbf{0.0407} \\
Ours                & \textbf{0.0357} & 0.0413          & 0.0422          & \textbf{0.0280} & 0.0598          & 0.0414          \\ \hline
\multicolumn{7}{c}{$SAM$ (degree)}                                                                                              \\ \hline
                    & BGU\_00257      & BGU\_00259      & BGU\_00261      & BGU\_00263      & BGU\_00265      & Average         \\
Interpolation       & 3.9620          & 3.0304          & 4.2962          & 3.1900          & 3.9281          & 3.6813          \\
Arad $et$ $al.$     & 4.2667          & 3.7279          & 3.4726          & 3.3912          & 3.3699          & 3.6457          \\
A+                  & 3.2952          & 3.5812          & 3.2952          & 3.0256          & 3.2952          & 3.2985          \\
Galliani $et$ $al.$ & 1.4725          & 1.5013          & \textbf{1.4802} & 1.4844          & \textbf{1.8229} & 1.5523          \\
Ours                & \textbf{1.3305} & \textbf{1.2458} & 1.7197          & \textbf{1.1360} & 1.9046          & \textbf{1.4673} \\ \hline
\end{tabular} }
\end{table}

\begin{figure*}[ht]
\begin{center}
\subfigure{\includegraphics[height=1.5in, width=1.5in]{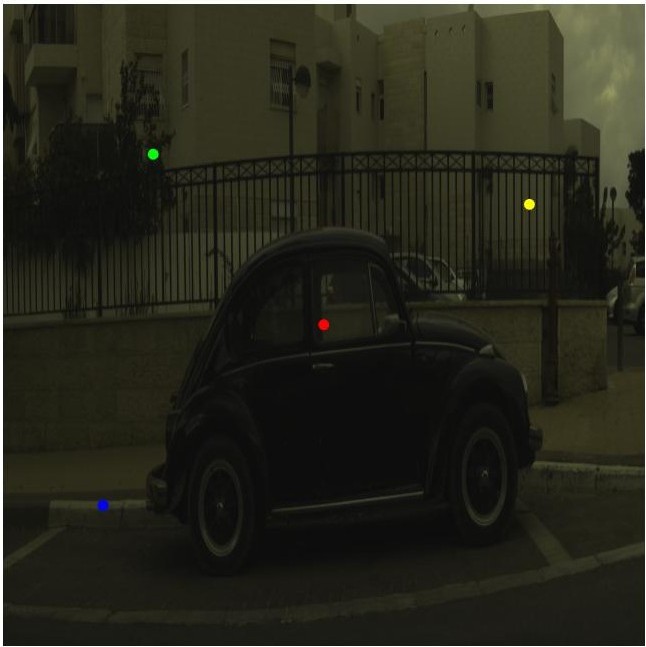}}
\subfigure{\includegraphics[height=1.5in, width=1.5in]{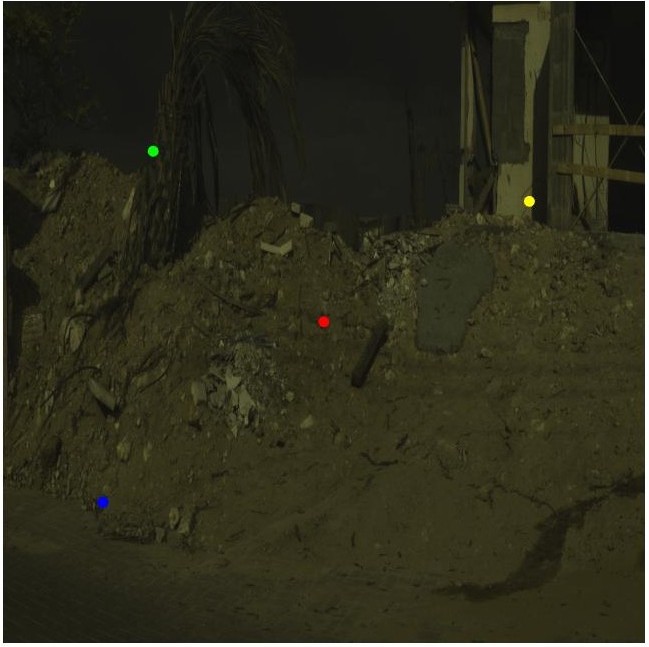}}
\subfigure{\includegraphics[height=1.5in, width=1.5in]{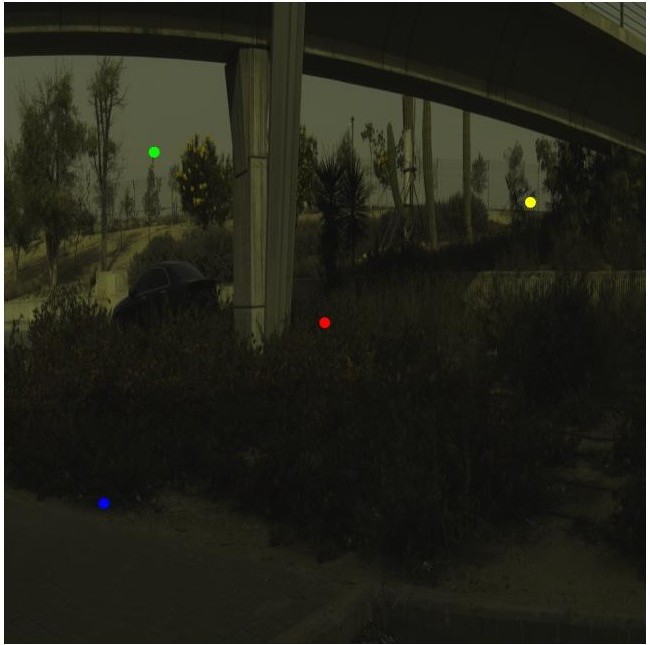}}
\subfigure{\includegraphics[height=1.5in, width=1.5in]{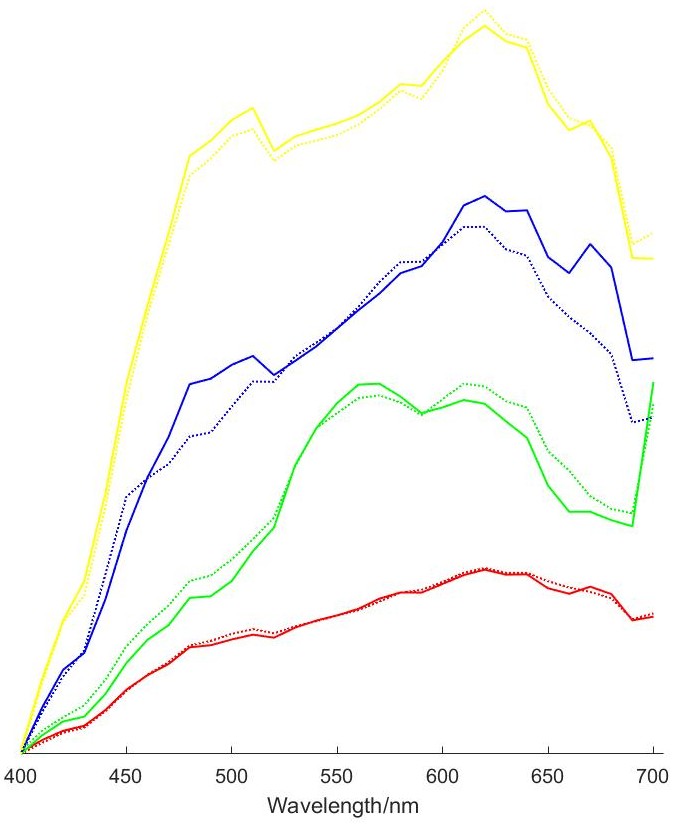}}
\subfigure{\includegraphics[height=1.5in, width=1.5in]{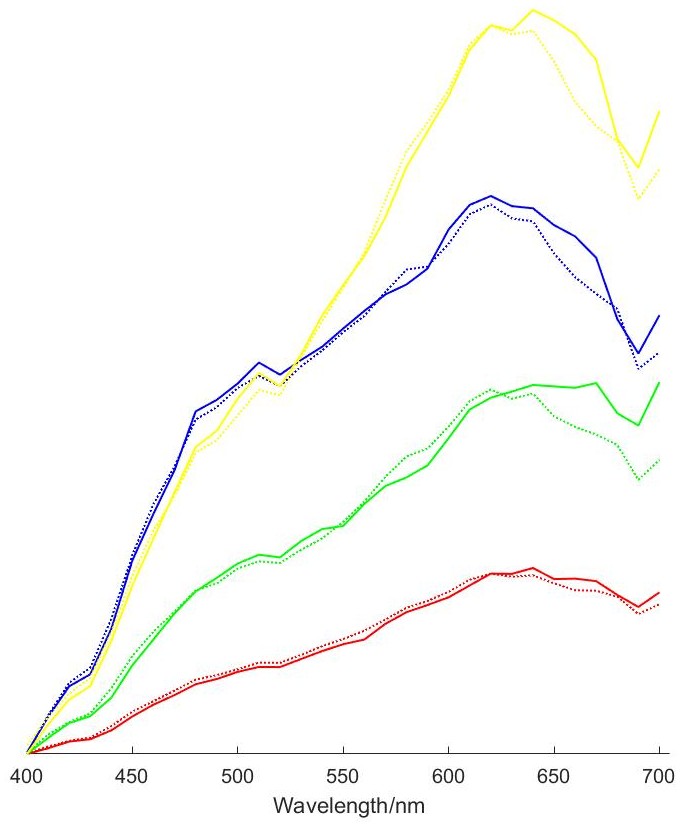}}
\subfigure{\includegraphics[height=1.5in, width=1.5in]{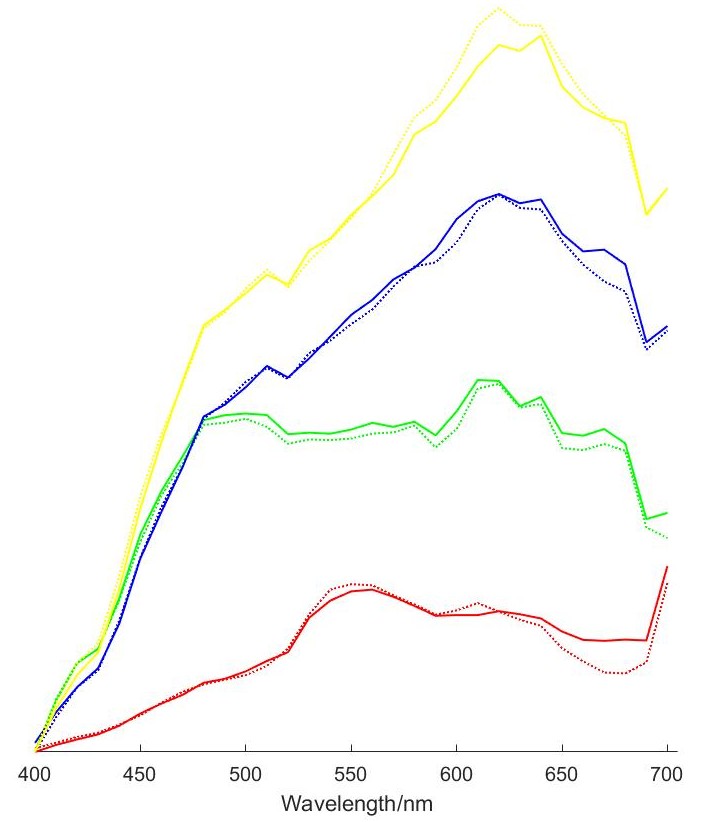}}
\end{center}
\caption{Sample results of spectral reconstruction by our method. Top line: RGB rendition. Bottom line: groundtruth (solid) amd reconstructed (dashed) spectral response of four pixels identified by the dots in RGB images.}
\label{fig3}
\end{figure*}

\subsection{Comparison Methods \& Implementation Details}

To demonstrate the effectiveness of the proposed method, we compare it with four spectral super-resolution methods, including spline interpolation, the sparse recovery method in~\cite{Reference20} (Arad $et$ $al.$), A+~\cite{Reference23}, and the deep learning method in~\cite{Reference30} (Galliani $et$ $al.$). \cite{Reference20,Reference23} are implemented by the codes released by the authors. Since there is no code released for~\cite{Reference30}, we reimplement it in this study. In the following, we will give the implementation details of each method.

\textbf{Spline interpolation} The interpolation algorithm serves as the most primitive baseline in this study. Specifically, for each RGB pixel $\bm{p}_{l} =  \big( r,g,b \big)$, we use spline interpolation to upsample it and obtain a $31$-dimensional spectrum ($\bm{p}_{h}$). According to the visible spectrum\footnote[1]{\url{http://www.gamonline.com/catalog/colortheory/visible.php}}, the $r$, $g$, $b$ values of an RGB pixel are assigned to $700nm$, $550nm$, and $450nm$, respectively.

\textbf{Arad $et$ $al.$ and A+} The low spectral resolution image is assumed to be a directly downsampled version of the corresponding hyperspectral image using some specific linear projection matrix. In \cite{Reference20,Reference23} this matrix is required to be perfectly known. In our experiments, we fit the projection matrix using training data with conventional linear regression.

\begin{figure*}[ht]
\begin{center}
\subfigure[Training curve]{\includegraphics[height=1.2in, width=1.5in]{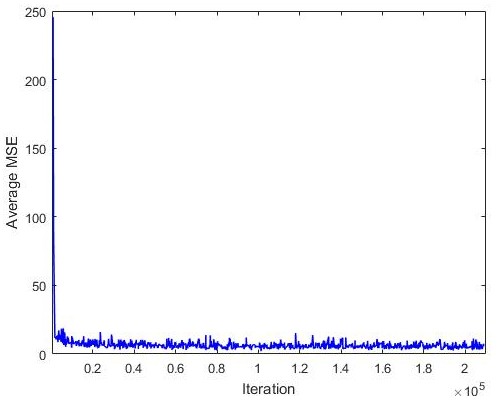}}
\subfigure[$RMSE_{1}$ test curve]{\includegraphics[height=1.2in, width=1.5in]{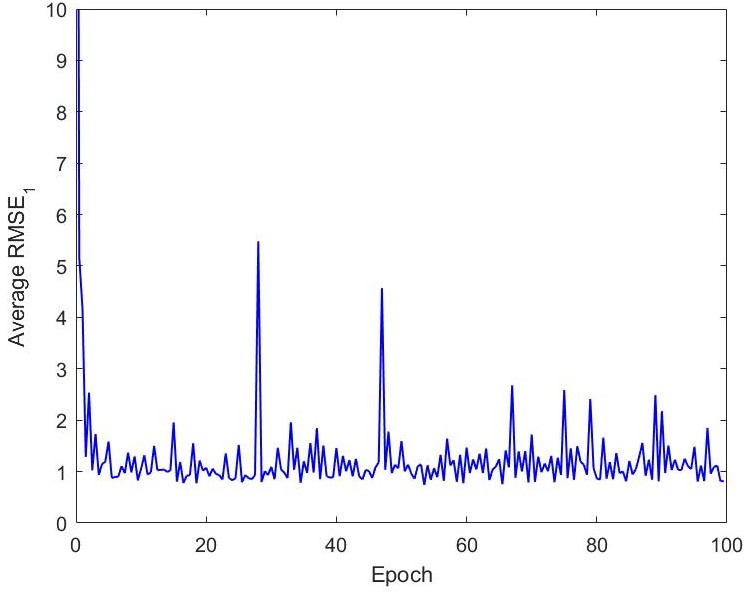}}
\subfigure[$RMSE_{2}$ test curve]{\includegraphics[height=1.2in, width=1.5in]{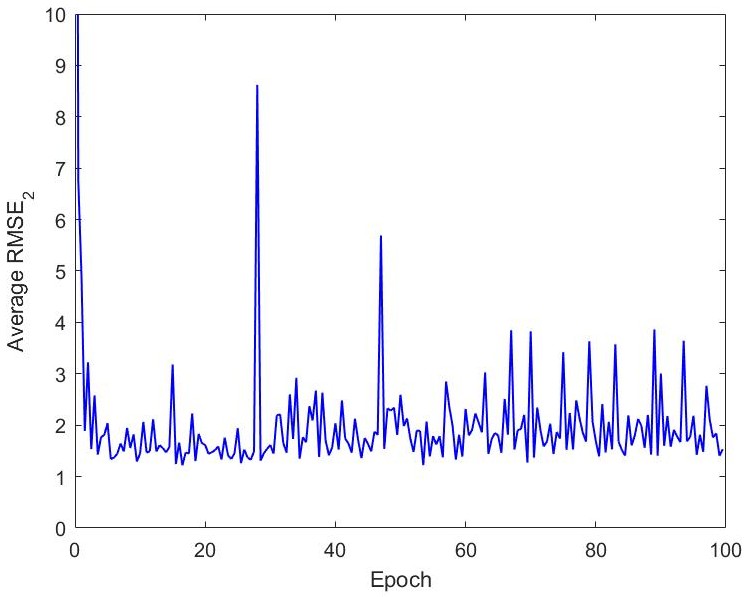}}
\subfigure[$rRMSE_{1}$ test curve]{\includegraphics[height=1.2in, width=1.5in]{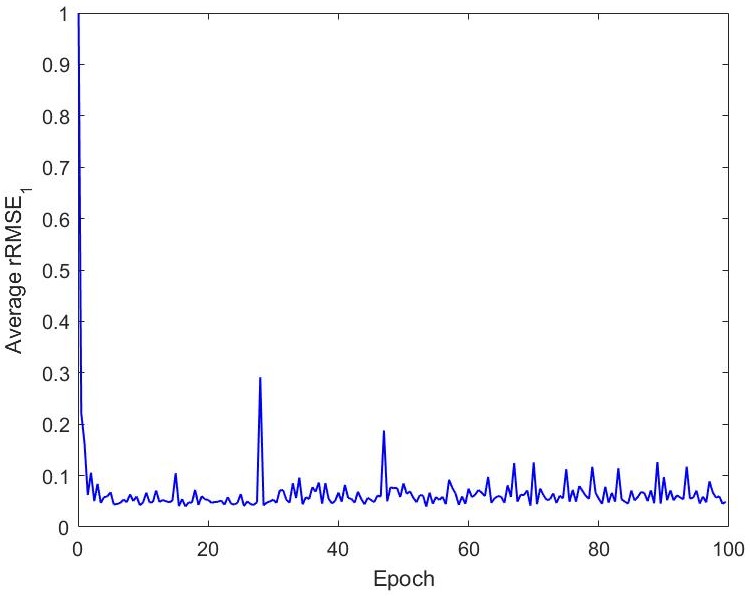}}
\subfigure[$rRMSE_{2}$ test curve]{\includegraphics[height=1.2in, width=1.5in]{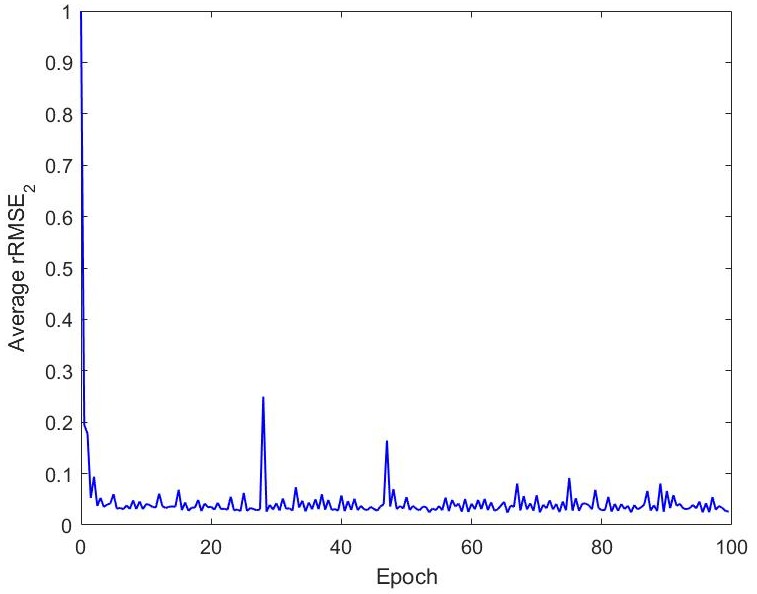}}
\subfigure[$SAM$ test curve]{\includegraphics[height=1.2in, width=1.5in]{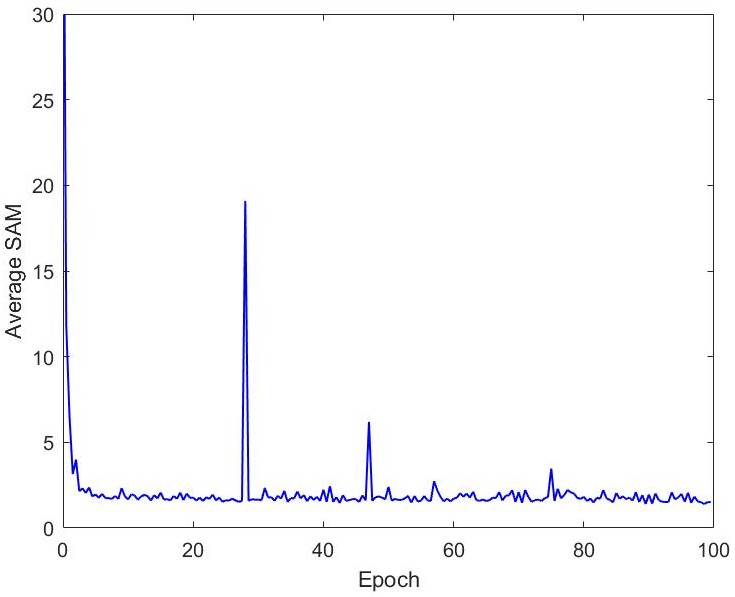}}
\end{center}
\caption{Training and test curves.}
\label{fig5}
\end{figure*}

\textbf{Galliani $et$ $al.$ and our method} We experimentally find the optimal set of hyper-parameters for both methods. $50\%$ dropout is applied to Galliani $et$ $al.$, while our method utilizes $20\%$ dropout rate. All the leaky ReLU activation functions are applied with a negative slope of 0.2. We train the networks for 100 epochs using Adam optimizer with $10^{-6}$ regularization. Weight initialization and learning rate vary for different methods. For Galliani $et$ $al.$, the weights are initialized via HeUniform~\cite{Reference48}, and the learning rate is set to $2 \times 10^{-3}$ for the first 50 epochs, decayed to $2 \times 10^{-4}$ for the next 50 epochs. As for our method, we use HeNormal initialization~\cite{Reference48}. The initial learning rate is $5 \times 10^{-5}$ and is multiplied by 0.93 every 10 epochs. We perform data augmentation by extracting patches of size $64 \times 64$ with a stride of 40 pixels from training data. The total amount of training samples is over $267,000$. At the test phase, we directly feed the whole image to the network and get the estimated hyperspectral image in one single forward pass.

\subsection{Evaluation Metrics}

To quantitatively evaluate the performance of the proposed method, we adopt the following two categories of evaluation metrics.

\textbf{Pixel-level reconstruction error} We follow~\cite{Reference23} to use absolute and relative root-mean-square error (RMSE and rRMSE) as quantitative measurements for reconstruction accuracy. Let $I_{h}^{(i)}$ and $I_{e}^{(i)}$ denote the $i$th element of the real and estimated hyperspectral images, $\bar{I}_{h}$ is the average of $I_{h}$, and $n$ is the total number of elements in one hyperspectral image. There are two formulas for RMSE and rRMSE respectively.

\begin{displaymath}
\begin{aligned}
RMSE_{1} = \frac{1}{n}\sum_{i=1}^{n}\sqrt{\left(I_{h}^{(i)}-I_{e}^{(i)}\right)^{2}} & \qquad
RMSE_{2} = \sqrt{\frac{1}{n}\sum_{i=1}^{n}\left(I_{h}^{(i)}-I_{e}^{(i)}\right)^{2}} \\
rRMSE_{1} = \frac{1}{n}\sum_{i=1}^{n}\frac{\sqrt{\left(I_{h}^{(i)}-I_{e}^{(i)}\right)^{2}}}{I_{h}^{(i)}} & \qquad
rRMSE_{2} = \sqrt{\frac{1}{n}\sum_{i=1}^{n}\frac{\left(I_{h}^{(i)}-I_{e}^{(i)}\right)^{2}}{\bar{I}_{h}^{2}}}
\end{aligned}
\end{displaymath}

\textbf{Spectral similarity} Since the key for spectral super-resolution is to reconstruct the spectra, we also use spectral angle mapper ($SAM$) to evaluate the performance of different methods. $SAM$ calculates the average spectral angle between the spectra of real and estimated hyperspectral images. Let $\bm{p}_{h}^{(j)}, \bm{p}_{e}^{(j)} \epsilon \ \mathbb{R}^{C}$ represents the spectra of the $j$th hyperspectral pixel in real and estimated hyperspectral images ($C$ is the number of bands), and $m$ is the total number of pixels within an image. The $SAM$ value can be computed as follows.

\begin{displaymath}
SAM = \frac{1}{m}cos^{-1}\left(\sum_{j=1}^{m}\frac{(\bm{p}_{h}^{(j)})^{T} \bm{\cdot} \bm{p}_{e}^{(j)}}{\left \| \bm{p}_{h}^{(j)} \right \|_{2} \bm{\cdot} \left \| \bm{p}_{e}^{(j)} \right \|_{2}} \right)
\end{displaymath}

\subsection{Experimental Results}

\textbf{Convergence Analysis} We plot the curve of $MSE$ loss on the training set and the curves of five evaluation metrics computed on the test set in Figure~\ref{fig5}. It can be seen that both the training loss and the value of metrics gradually decrease and ultimately converge with the proceeding of the training. This demonstrates that the proposed multi-scale convolution neural network converges well.

\textbf{Quantitative Results} Table~\ref{tab4} provides the quantitative results of our method and all baseline methods. It can be seen that our model outperforms all competitors with regards to $RMSE_{1}$ and $rRMSE_{1}$, and produces comparable results to Galliani $et$ $al.$ on $RMSE_{2}$ and $rRMSE_{2}$. More importantly, our method surpasses all the others with respect to spectral angle mapper. This clearly proves that our model reconstructs spectra more accurately than other competitors. It is worth pointing out that reconstruction error (absolute and relative $RMSE$) is not necessarily positively correlated with spectral angle mapper ($SAM$). For example, when the pixels of an image are shuffled, $RMSE$ and $rRMSE$ will remain the same, while $SAM$ will change completely. According to the results in Table~\ref{tab4}, we can find that our finely designed network enhances spectral super-resolution from both aspects, $viz.$, yielding better results on both average root-mean-square error and spectral angle similarity.

\begin{figure*}[!t]
\begin{center}
\subfigure{\includegraphics[height=1.0in, width=1.0in]{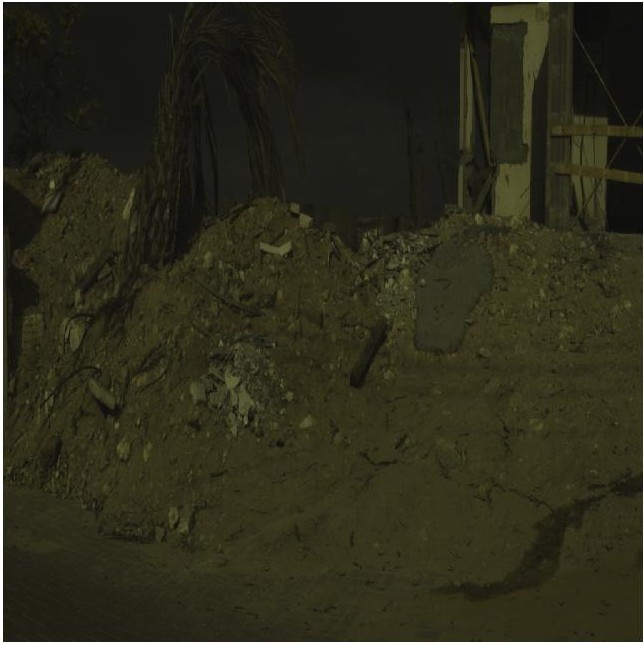}}
\subfigure{\includegraphics[height=1.0in, width=1.0in]{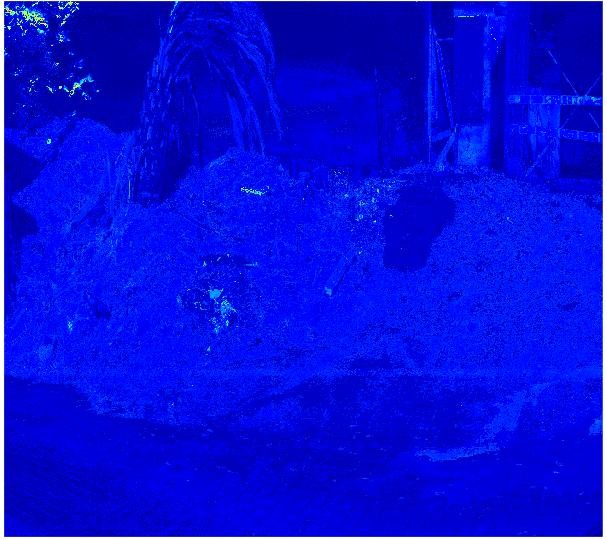}}
\subfigure{\includegraphics[height=1.0in, width=1.0in]{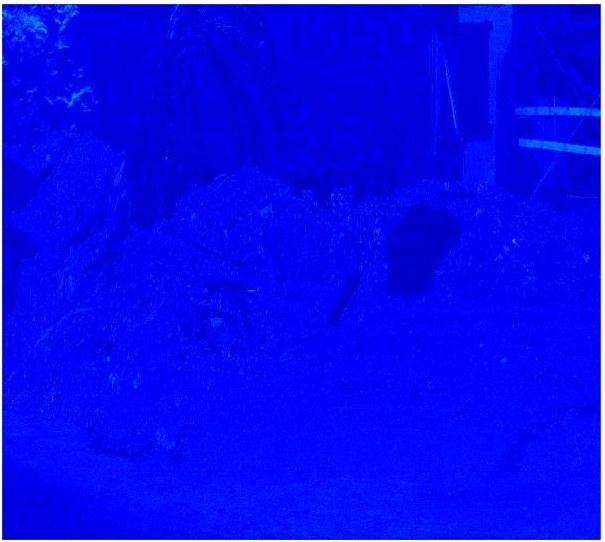}}
\subfigure{\includegraphics[height=1.0in, width=1.0in]{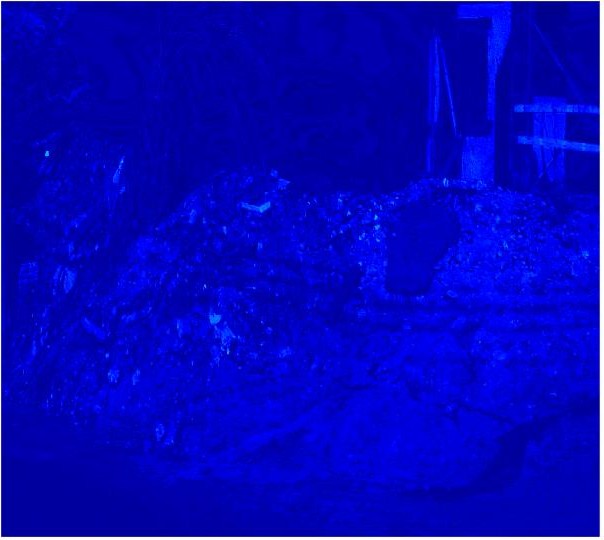}}
\subfigure{\includegraphics[height=1.0in, width=0.15in]{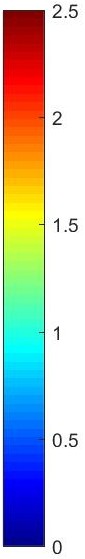}}
\subfigure{\includegraphics[height=1.0in, width=1.0in]{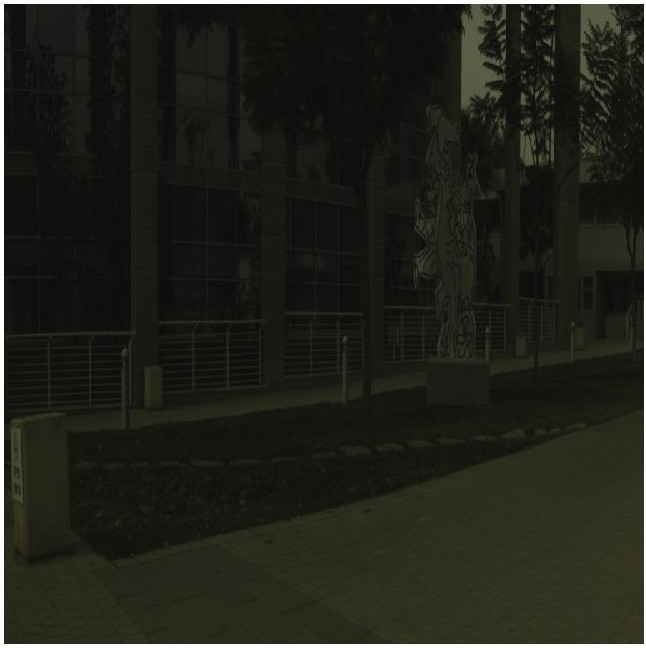}}
\subfigure{\includegraphics[height=1.0in, width=1.0in]{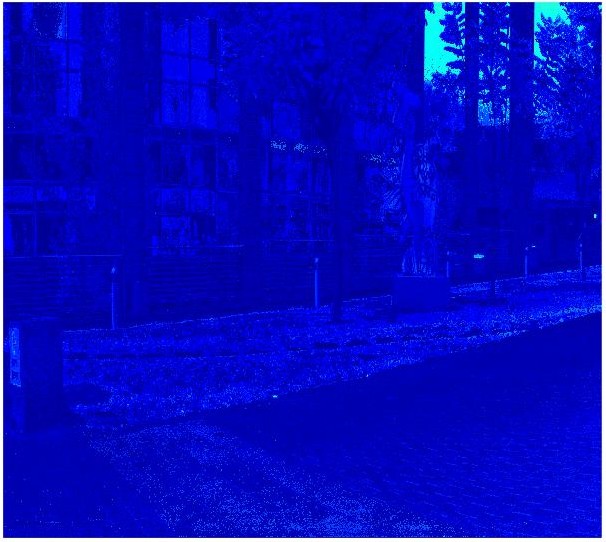}}
\subfigure{\includegraphics[height=1.0in, width=1.0in]{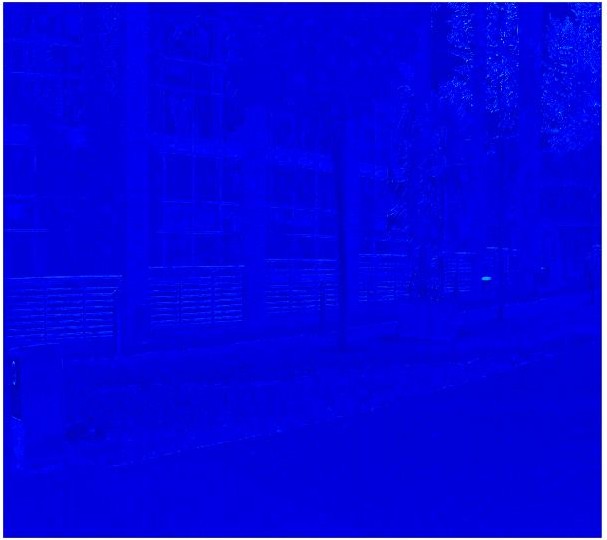}}
\subfigure{\includegraphics[height=1.0in, width=1.0in]{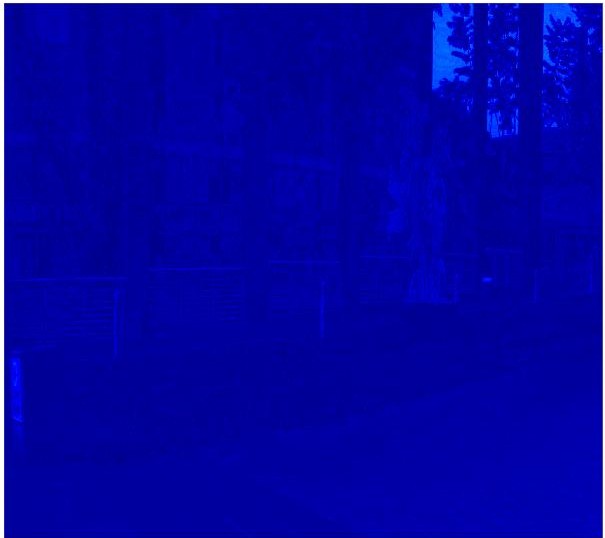}}
\subfigure{\includegraphics[height=1.0in, width=0.15in]{fig2_Colorbar.jpg}}
\subfigure{\includegraphics[height=1.0in, width=1.0in]{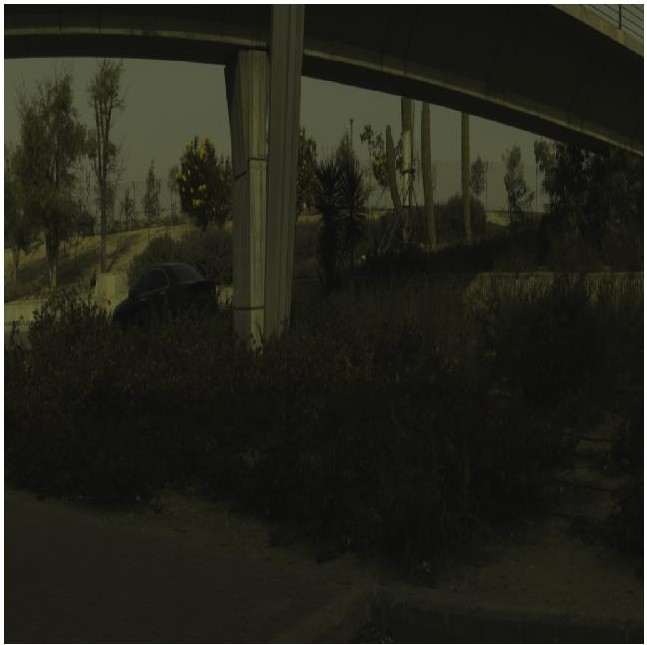}}
\subfigure{\includegraphics[height=1.0in, width=1.0in]{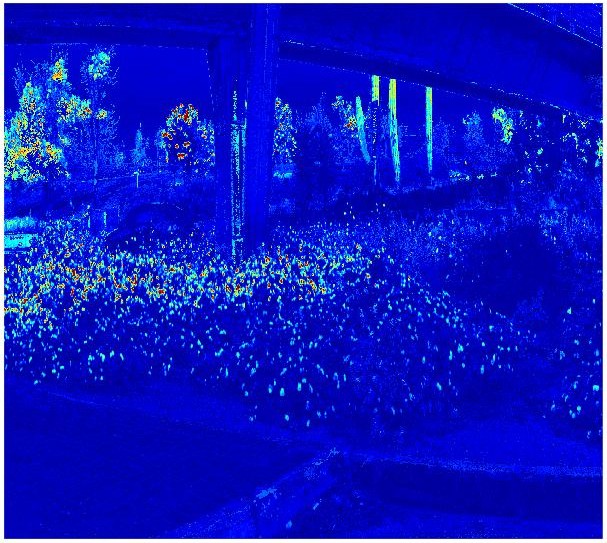}}
\subfigure{\includegraphics[height=1.0in, width=1.0in]{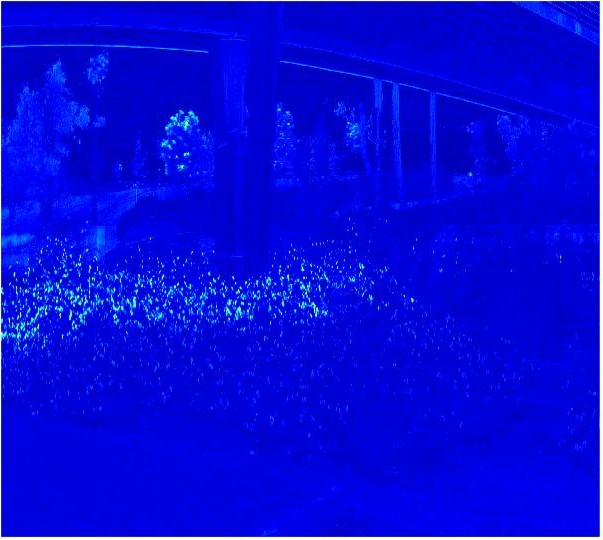}}
\subfigure{\includegraphics[height=1.0in, width=1.0in]{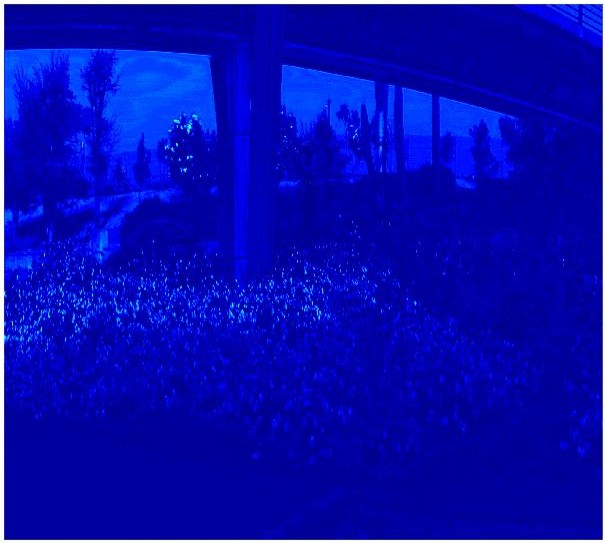}}
\subfigure{\includegraphics[height=1.0in, width=0.15in]{fig2_Colorbar.jpg}}
\end{center}
\caption{Visualization of absolute reconstruction error. From left to right: RGB rendition, A+, Galliani $et$ $al.$, and our method}
\label{fig2}
\end{figure*}

\textbf{Visual Results} To further clarify the superiority in reconstruction accuracy. We show the absolute reconstruction error of test images in Figure~\ref{fig2}. The error is summarized over all bands of the hyperspectral image. Since A+ outperforms Arad $et$ $al.$ in terms of any evaluation metric, we use A+ to represent the sparse coding methods. It can be seen that our method yields smoother reconstructed images as well as lower reconstruction error than other competitors.

In addition, we randomly choose three test images and plot the real and reconstructed spectra for four pixels in Figure~\ref{fig3} to further demonstrate the effectiveness of the proposed method in spectrum reconstruction. It can be seen that only slight difference exists between the reconstructed spectra and the ground truth.

According to these results above, we can conclude that the proposed method is effective in spectral super-resolution and outperforms several state-of-the-art competitors.

\section{Conclusion}

In this study, we show that leveraging both the local and non-local information of input images is essential for the accurate spectral reconstruction. Following this idea, we design a novel multi-scale convolutional neural network, which employs a symmetrically cascaded downsampling-upsampling architecture to jointly encode the local and non-local image information for spectral reconstruction. With extensive experiments on a large hyperspectral images dataset, the proposed method clearly outperforms several state-of-the-art methods in terms of reconstruction accuracy and spectral similarity.

\section{Acknowledgement}

This work was supported in part by the National Natural Science Foundation of China (No. 61671385, 61571354), Natural Science Basis Research Plan in Shaanxi Province of China (No. 2017JM6021, 2017JM6001) and China Postdoctoral Science Foundation under Grant (No. 158201).

%
%
%


\bibliographystyle{splncs04}
\bibliography{Bibliography}

%

\end{document}